\documentclass[10pt,twocolumn,letterpaper]{article}

\usepackage{cvpr}
\usepackage{times}
\usepackage{epsfig}
\usepackage{graphicx}
\usepackage{amsmath}
\usepackage{amssymb}
\usepackage{balance}
\usepackage{lipsum}
\usepackage{booktabs}
\usepackage{array,multirow}
\usepackage{xcolor}
\usepackage[british,american]{babel}
\usepackage[pagebackref=true,breaklinks=true,letterpaper=true,colorlinks,bookmarks=false]{hyperref}

\newcommand{\figref}[1]{\mbox{Fig. \ref{#1}}}
\newcommand{\tblref}[1]{\mbox{Table \ref{#1}}}
\newcommand{\secref}[1]{\mbox{Sec. \ref{#1}}}
\renewcommand{\eqref}[1]{\mbox{Eq. \ref{#1}}}

\newcommand{\ours}{\mbox{{CAAP}~}}
\newcommand{\sps}[1]{\mathcal{#1}} 
\usepackage[font=small]{subcaption}
\let\oldparagraph\paragraph
\renewcommand{\paragraph}[1]{\vspace{-0.30cm} \oldparagraph{#1}}

\newcommand{\figvspace}{\vspace{-0.3cm}}
\newcommand{\figlblvspace}{\vspace{-0.2cm}}
\newcommand{\eqtopvspace}{\vspace{-0.1cm}}
\newcommand{\eqbottomvspace}{\vspace{-0.0cm}}
\newcommand{\tabvspace}{\vspace{-0.2cm}}
\newcommand{\tablblvspace}{\vspace{-0.2cm}}
\newcommand{\reffontsize}{\small} 
\newcommand{\todos}[1]{} 
\cvprfinalcopy 
\def\cvprPaperID{121} 

\ifcvprfinal\fi
\begin{document}

\title{Recovering the Missing Link: Predicting Class-Attribute Associations\\ for Unsupervised Zero-Shot Learning}

\author{Ziad Al-Halah \hspace{2cm} Makarand Tapaswi \hspace{2cm} Rainer Stiefelhagen \\
Karlsruhe Institute of Technology, 76131 Karlsruhe, Germany\\
{\tt\small \{ziad.al-halah, makarand.tapaswi, rainer.stiefelhagen\}@kit.edu}
}

\maketitle

\begin{abstract}
\makeatletter{}
Collecting training images for all visual categories is not only expensive but also impractical.
Zero-shot learning (ZSL), especially using attributes, offers a pragmatic solution to this problem.
However, at test time most attribute-based methods require a full description of attribute associations for each unseen class.
Providing these associations is time consuming and often requires domain specific knowledge.
In this work, we aim to carry out attribute-based zero-shot classification in an unsupervised manner.
We propose an approach to learn relations that couples class embeddings with their corresponding attributes.
Given only the name of an unseen class, the learned relationship model is used to automatically predict the class-attribute associations.
Furthermore, our model facilitates transferring attributes across data sets without additional effort.
Integrating knowledge from multiple sources results in a significant additional improvement in performance.
We evaluate on two public data sets: Animals with Attributes and aPascal/aYahoo.
Our approach outperforms state-of-the-art methods in both predicting class-attribute associations and unsupervised ZSL by a large margin.

\end{abstract}

\section{Introduction}
\makeatletter{}
\begin{figure}[!t]
\centering
\includegraphics[width=\linewidth]{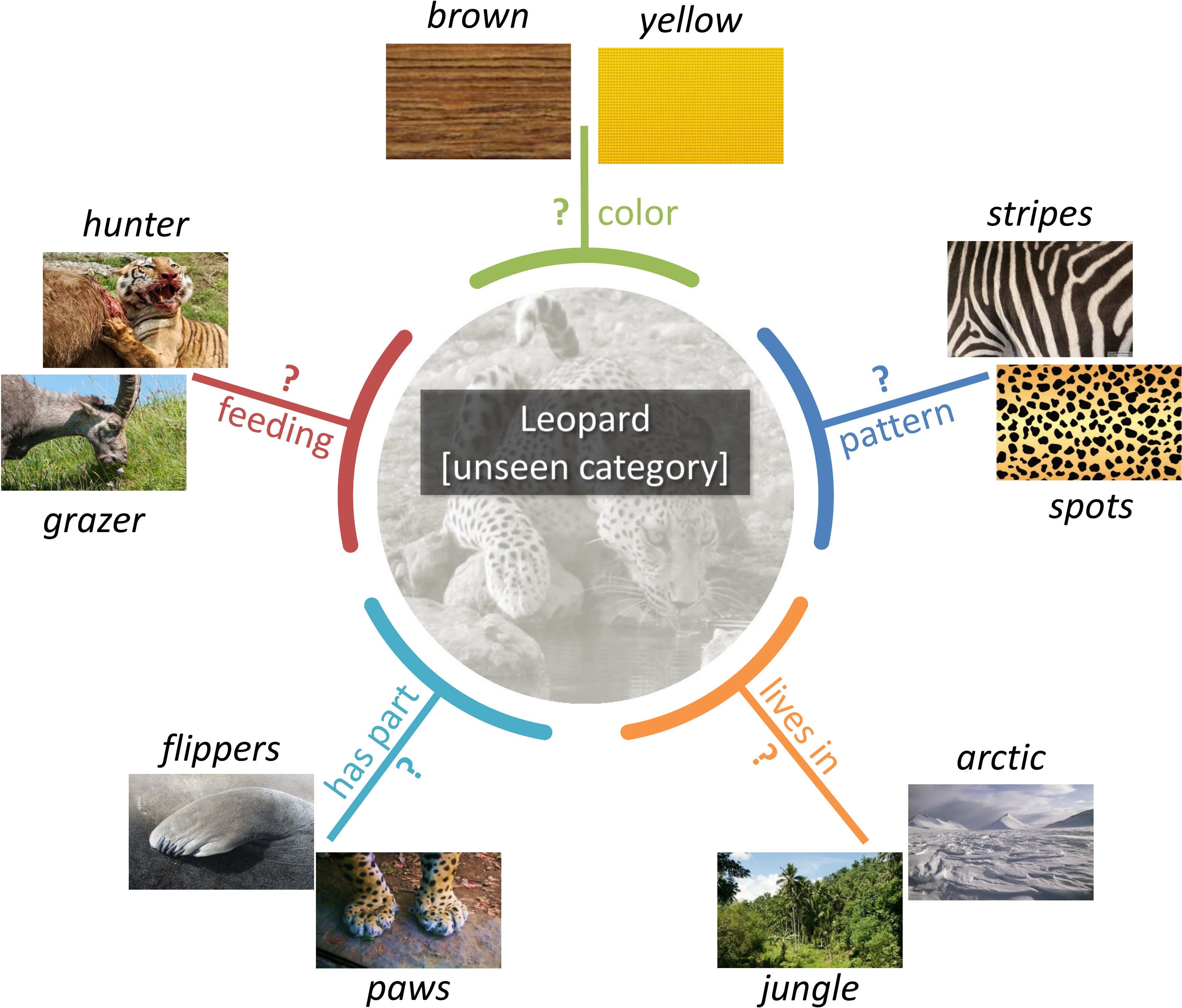}
\figlblvspace
\caption{Given only the name of an unseen category, here \emph{Leopard}, our method automatically predicts the list of attributes (\eg~yellow, spots) associated with the class through relationships (\eg~has\_color, has\_pattern).
These predicted associations are leveraged to build category classifiers for zero-shot learning.}
\label{fig:firstpage}
\figvspace
\end{figure}

\makeatletter{}

Large-scale object classification and visual recognition have seen rising interest in the recent years.
Data sets such as ImageNet~\cite{Deng2009} have helped scale up the number of classes represented in tasks such as object classification or detection.
Many methods based on deep convolutional neural networks~\cite{Krizhevsky2012,Szegedy2014} have been developed recently to leverage the power of millions of training images distributed among thousands of classes.
However, building a large data set, especially collecting large number of training images is very challenging and nevertheless this ends up representing only a fraction of the real visual world~\cite{Biederman1987}.

Transfer learning is a practical solution to bridge this gap as it allows to leverage knowledge and experience obtained from existing data to new domains.
Specifically for object classification, knowledge from object categories which have labeled image samples can be transferred to new unseen categories which do not have training images.
This task is referred to as zero-shot learning (ZSL).

There exist many directions in the literature to perform ZSL.
These primarily differ in the type of knowledge source they tap in order to establish the connection between the unseen classes and the available visual information~\cite{Bart2005a,Lampert2009,Rohrbach2011}.
Among these directions, attribute-based knowledge transfer shows impressive performance~\cite{Al-Halah2015,Lampert2009,Liu2011}.
By learning an intermediate layer of semantic attributes (\eg~colors or shapes), a novel class is then described with a subset of these attributes and its model is constructed accordingly based on the respective attribute classifiers.

A major drawback of attribute-based approaches is that user supervision is needed to provide the description for each novel class.
For example, for the new class ``leopard'' the user needs to describe it with a set of visual attributes in order to establish the semantic link with the learned visual vocabulary (\eg~the leopard has part paws, it exhibits a spotted pattern but does not live in water).
This amounts to providing manual class-attribute associations in the range of tens~\cite{Farhadi2009,Lampert2009} to hundreds~\cite{Wah2011} of attributes for each new category.
This is not only time consuming but often also requires domain-specific or expert knowledge~\cite{Lampert2009,Wah2011} that the user is unlikely to have.
It is more convenient and intuitive for the user to provide just the name of the unseen class rather than a lengthy description.

Our goal is to remove this need for attribute supervision when performing zero-shot classification.
We aim to automatically link a novel category with the visual vocabulary and predict its attribute association without user intervention.
Thereby, we answer questions such as: Does the leopard live in the jungle? Does it have a striped pattern? (see~\figref{fig:firstpage}).
To this end, we propose a novel approach that learns semantic relations and automatically associates an unseen class with our visual vocabulary (\ie~the attributes) based solely on the class name.
Using the predicted relations, we are able to construct a classifier of the novel class and conduct unsupervised zero-shot classification.
Moreover, we demonstrate that our model is even able to automatically transfer the visual vocabulary itself across data sets which results in significant performance improvements at no additional cost.
We demonstrate the effectiveness of such a model against state-of-the-art via extensive experiments.

\section{Related work}
\makeatletter{}

In ZSL the set of train and test classes are disjoint.
That is, while we have many labeled samples of the train classes to learn a visual model, we have never observed examples of the test class (a.k.a. unseen class).
In order to construct a visual model for the unseen class, we first need to establish its relation to the visual knowledge that is obtained from the training data.
One of the prominent approaches in the literature is attribute-based ZSL.
Attributes describe visual aspects of the object, like its shape, texture and parts~\cite{Ferrari2008}.
Hence, the recognition paradigm is shifted from labeling to describing~\cite{Chen2015,Sadovnik2013,Saleh2013}.
In particular, attributes act as an intermediate semantic representation that can be easily transferred and shared with new visual concepts~\cite{Escorcia2015,Farhadi2009,Lampert2009}.
In ZSL, attributes have been used either directly~\cite{Farhadi2009,Lampert2009,Liu2011}, guided by hierarchical information~\cite{Al-Halah2015}, or in transductive settings~\cite{Fu2014,Rohrbach2013}.

However, most attribute-based ZSL approaches rely on the underlying assumption that for an unseen class the complete information about attribute associations are manually defined~\cite{Farhadi2009} or imported from expert-based knowledge sources~\cite{Lampert2009,Wah2011}.
This is a hindering assumption since the common user is unlikely to have such a knowledge or is simply unwilling to manually set hundreds of associations for each new category.

Towards simplifying the required user involvement, given an unseen class \cite{Yu2013} reduces the level of user intervention by asking the operator to select the most similar seen classes and then inferring its expected attributes.
\cite{Mensink2014,Rohrbach2010} go a step further and propose an unsupervised approach to automatically learn the class-attribute association strength by using text-based semantic relatedness measures and co-occurrence statistics obtained from web-search hit counts.
However, as web data is noisy, class and attribute terms can appear in documents in different contexts which are not necessarily related to the original attribute relation we seek.
We demonstrate in this work, that the class-attribute relations are complex and it is hard to model them by simple statistics of co-occurrence.

In an effort to circumvent the need for manually defined associations, \cite{Ba2015,Elhoseiny2013} propose to extract pseudo attributes from Wikipedia articles using TF-IDF based embeddings to predict the visual classifier of an unseen class.
In theory, an article can be extracted automatically by searching for a matching title to the class name.
However, in practice manual intervention is needed when there is no exact match or the article is titled with a synonym or the scientific name of the category as reported by~\cite{Elhoseiny2013}.

In a different direction, unsupervised ZSL can be conducted by exploiting lexical hierarchies.
For example, \cite{Rohrbach2011} uses WordNet~\cite{Miller1995} to find a set of ancestor categories of the novel class and transfer their visual models accordingly.
Likewise, \cite{Al-Halah2015} uses the hierarchy to transfer the attribute associations of an unseen class from its seen parent in the ontology.
In~\cite{Akata2015}, WordNet is used to capture semantic similarity among classes in a structured joint embedding framework.
However, categories that are close to each other in the graph (\eg~siblings) often exhibit similar properties to their ancestors making it hard to discriminate among them.
Moreover, ontologies like WordNet are not complete.
Many classes (\eg fine-grained) are not present in the hierarchy.

Recently, \cite{Frome2013,Socher2013} proposed to learn a direct embedding of visual features into the semantic word space of categories.
They leverage a powerful neural word embedding~\cite{Huang2012,Mikolov2013} that is trained on a large text corpus, and learn a mapping from the space of visual features to the word representation.
At test time, they predict an image class by looking for the nearest category embedding to the one estimated by the neural network.
\cite{Frome2013}~shows impressive results of this approach for large-scale ZSL.
\cite{Norouzi2014} improves upon \cite{Frome2013} by considering a convex combination of word embeddings weighted by classifiers confidences to estimate the unseen  class embedding.
However, we show in our evaluation that such word embedding approaches are less discriminative than their attribute-based counterpart.

We propose an approach that goes beyond using web statistics, predefined ontologies and word embedding estimation.
We provide an automatic framework to learn complex class-attribute relations and effectively transfer knowledge across domains for unsupervised zero-shot learning. 

\section{Approach}
\makeatletter{}
\makeatletter{}
\begin{figure*}[!t]
\centering
\begin{subfigure}[b]{0.32\linewidth}
    \includegraphics[width=\linewidth]{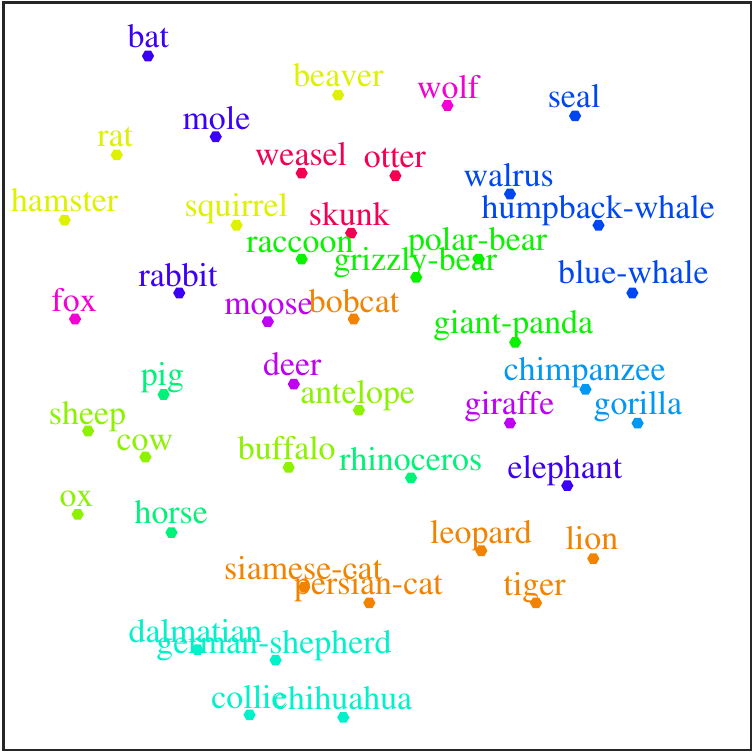}
    \caption{Class vectors}\label{fig:tsne_cls}
\end{subfigure}\,
\begin{subfigure}[b]{0.32\linewidth}
    \includegraphics[width=\linewidth]{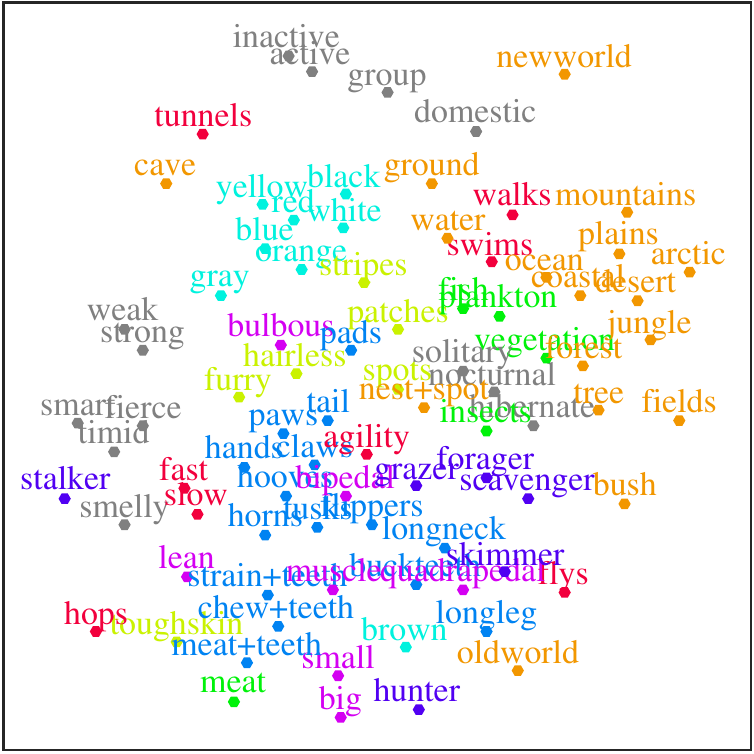}
    \caption{Attribute vectors}\label{fig:tsne_attr}
\end{subfigure}\,
\begin{subfigure}[b]{0.32\linewidth}
    \includegraphics[width=\linewidth]{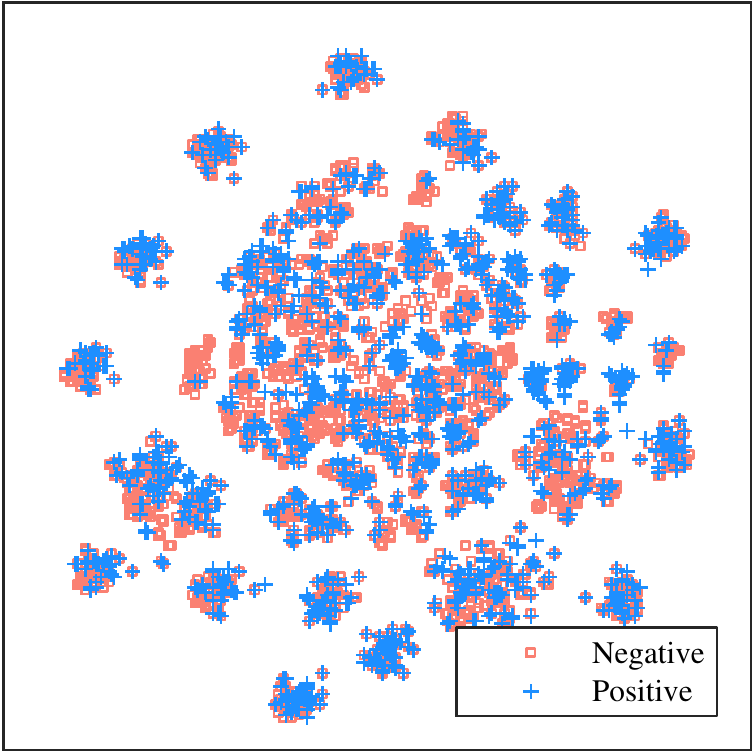}
    \caption{Class-Attribute displacement vectors}\label{fig:tsne_cls-attr}
\end{subfigure}
\figlblvspace
\caption{t-SNE representation of
(a) class embeddings: colors indicate similar classes based on the super category in the WordNet hierarchy (\eg~dalmatian, collie, and other dog breeds are all colored in cyan);
(b) attribute embeddings: colors indicate attributes which are grouped together to form class-attribute relations (\eg~has\_color relationship clusters all colors yellow, black, \etc which are represented in cyan); and
(c) class-attribute pair-wise displacement vectors (\eg $\mathrm{v(dolphin)}-\mathrm{v(ocean)}$) which show that encoding relationships using vector operations is a difficult task.
This figure is best viewed in color.}
\label{fig:tsne}
\figvspace
\end{figure*}
 
\begin{figure}[t]
\centering
\includegraphics[width=\linewidth]{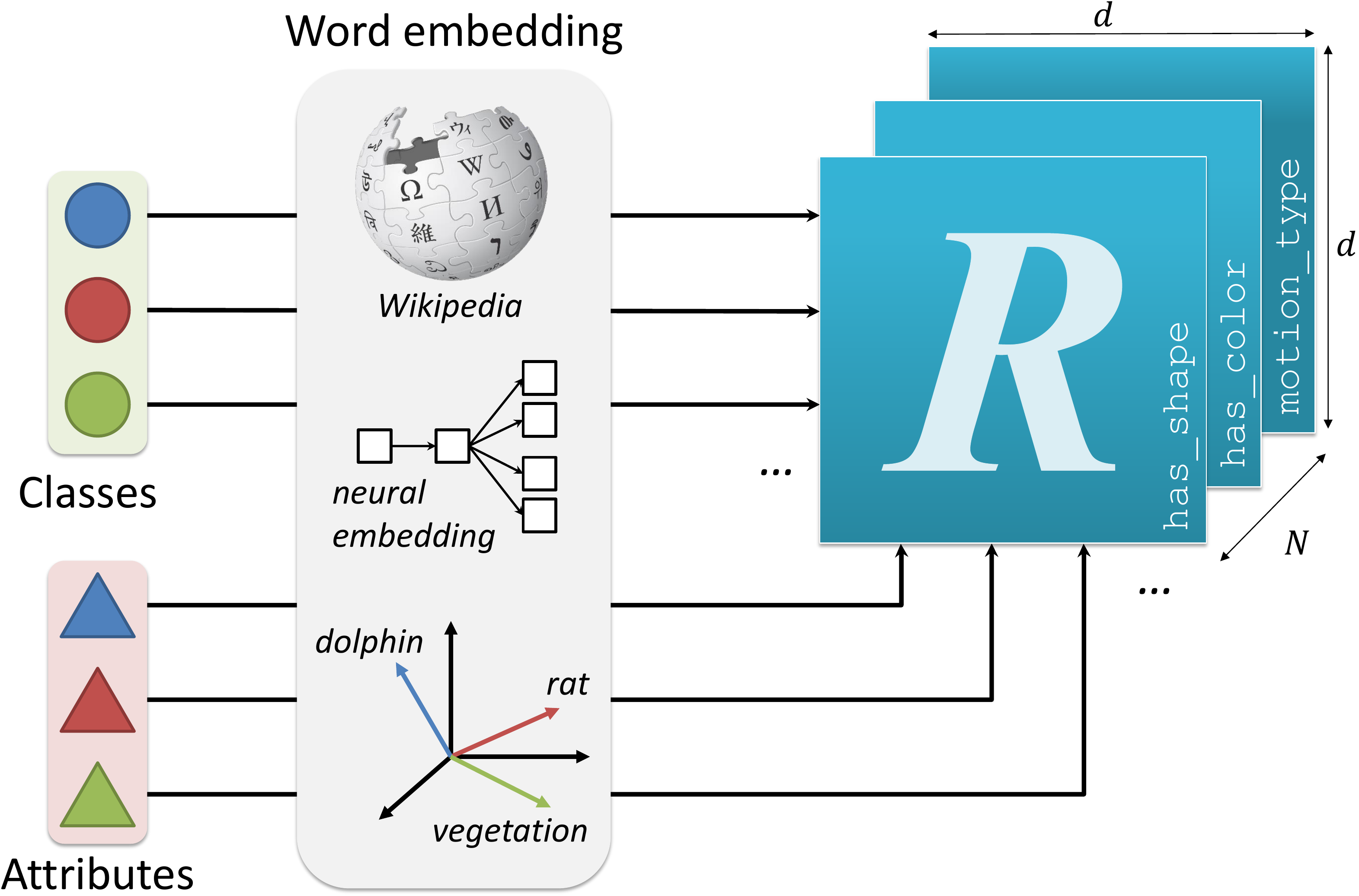}
\figlblvspace
\caption{Our model couples class and attribute embeddings using the tensor $\mathbf{R}$. Each slice $\mathbf{R}_j$ captures a relationship like has\_shape or motion\_type. The embeddings are obtained from a neural network trained on a large text corpus.}
\label{fig:model}
\figvspace
\end{figure}

We present an end-to-end approach to automatically predict class-attribute associations and use them for zero-shot classification.
We begin by (i) finding suitable vector representations for words and use the learned embedding as a way to mathematically relate class and attribute names.
These representations form the basis to model semantic relationships between classes and attributes.
(ii) We formulate the learning of these relations in a tensor factorization framework (see \figref{fig:model}) and offer key insights to adapt such a model to our problem.
Finally, (iii) for an unseen class we show how to predict the set of its most confident attribute associations and carry out zero-shot classification.
We start by defining the notation used throughout this paper.
\noindent \paragraph{Notation}
Let $\sps{C}=\{c_k\}^{K}_{k=1}$ be a set of seen categories that are described with a group of attributes $\sps{A}=\{a_m\}^{M}_{m=1}$.
The vector representation of a word is denoted by $\mathrm{v}(\cdot)$, and we use $\mathrm{v}(c_k)$ and $\mathrm{v}(a_m)$ for class $c_k$ and attribute $a_m$ respectively.
The categories and attributes are related by a set of relations $\sps{R}=\{r_j\}^{N}_{j=1}$ such that $r_j(c_k,a_m)=1$ if $c_k$ is connected to $a_m$ by relation $r_j$ and $0$ otherwise (\eg $\mathrm{has\_color(sky,blue)=1}$).
Given only the name of an unseen class $z \not \in \sps{C}$, our goal is to predict the attributes that are associated with the class (\eg $\mathrm{has\_color(\emph{whale},blue)=?}$) and conduct ZSL accordingly.

\subsection{Vector space embedding for words}\label{sec:app_word_embedding}
\makeatletter{}
In order to model the relations between classes and attributes, we require a suitable representation that transforms names to vectors while at the same time preserves the semantic connotations of the words.
Hereof, we use the skip-gram model presented by Mikolov~\etal~\cite{Mikolov2013} to learn vector space embeddings for words.
The skip-gram model is a neural network that learns vector representations for words that best help in predicting the surrounding words. Therefore, words that appear in a similar context (neighboring words) are represented with vectors that are close to each other in the embedding space.

\figref{fig:tsne} visualizes the obtained word vector representation for few classes and attributes in our data set using t-SNE~\cite{vanderMaaten2008}.
Even in such a low-dimension it is clear that classes related to each other appear closer.
This is evident for example from the group of dog breeds or feline in \figref{fig:tsne_cls}.
Similarly, we also see clusters in the attribute label space corresponding to colors, animal parts, and environment (see \figref{fig:tsne_attr}).

\paragraph{Relations in embedding space}
The skip-gram embeddings have gained popularity owing to their power in preserving useful linguistic patterns.
An empirical evaluation~\cite{Mikolov2013} shows that syntactic/semantic relations can be represented by simple vector operations in the word embedding space.
A great example is $\mathrm{v(king)} - \mathrm{v(man)} + \mathrm{v(woman)} \approx \mathrm{v(queen)}$, where $\mathrm{v(king)}$ is the embedding for ``king''.
In other words, the relation between ``king" and ``man" modeled by their displacement vector is similar to the displacement between ``queen" and ``woman".

However, modeling class-attribute relations by simple vector operations is inadequate.
\figref{fig:tsne_cls-attr} presents the t-SNE representation for \emph{displacement vectors} between each class-attribute pair (\eg~$\mathrm{v(sky)} - \mathrm{v(blue)}$).
We see that displacement vectors for both positive existing relations \emph{and} negative non-existing relations are inseparable. We empirically show in \secref{subsec:cls_attr_assoc} that class-attribute relations are more complicated and are not easily represented by simple vector operations.

To address this challenge we adopt a more sophisticated and comprehensive method to learn these relations while at the same time effectively exploit the powerful word embedding representation.
 
\makeatletter{}

\subsection{Learning class-attribute relations}\label{sec:app_relation_learning}
\makeatletter{}
We now model the complex relations between categories and their corresponding visual attributes.
Leveraging information based on these relations, we can predict the associations between a novel unseen class and our attribute vocabulary and build the corresponding ZSL classifier.

We propose to model the class-attribute relations using a tensor factorization approach~\cite{Nickel2012,Sutskever2009}. We represent the relations using a three dimensional tensor $\mathbf{R} \in \mathbb{R}^{d \times d \times N}$ where $d$ is the dimension of the word embedding and $N$ the number of relations (see \figref{fig:model}).
Each slice $\mathbf{R}_j \in \mathbb{R}^{d\times d}$ in the tensor models a relation $r_j$ (\eg~$\mathrm{has\_color}$) as a bilinear operator.
The likelihood of class $c_k$ being associated with attribute $a_m$ through relation $r_j$ is:
\begin{equation}
\eqtopvspace
p(r_j(c_k,a_m)) = \sigma(\mathrm{v}(c_k)^T\mathbf{R}_j\mathrm{v}(a_m)),
\eqbottomvspace
\end{equation}
where $\mathrm{v}(x)\in\mathbb{R}^d$ is the vector embedding of word $x$ and $\sigma(\cdot)$ is the logistic function.
We learn $\mathbf{R}$ by minimizing the negative log-likelihood of both positive ($\sps{P}$) and negative ($\sps{N}$) class-attribute associations for each slice $\mathbf{R_j}$: 
\newcommand{\sm}{\hspace{-12pt}}
\begin{equation}\label{eq:min_log}
\eqtopvspace
\begin{split}
\min_{\mathrm{v}(\sps{A}), \mathbf{R_j}} & 
- \sm \sum_{(j,k,m)\in \sps{P}} \sm \log(p(t_{k,m}^j \!=\! 1 ))\! -\! \sm \sum_{(j,k,m)\in \sps{N}}\sm \log(p(t_{k,m}^j \!=\! 0)),\\
\mathrm{where} \quad & t_{k,m}^j=r_j(c_k,a_m)
\end{split}
\eqbottomvspace
\end{equation}

Note that there are two key components in \eqref{eq:min_log}.
Firstly, we take advantage of the powerful representation of skip-gram and learn word embeddings on a large text corpus to initialize the embeddings of our class ($\mathrm{v}(\sps{C})$) and attribute ($\mathrm{v}(\sps{A})$) entities.
This gives our model the ability to generalize well to unseen classes and take advantage of the initial learned similarities among the attributes.
Secondly, in our case of zero-shot classification, the novel class name is not available during training and we have no information about how this unseen class is related with the visual attributes.
Consequently, we treat the set of categories as an \emph{open} set and fix their embedding $\mathrm{v}(\sps{C})$ to the one learned in~\secref{sec:app_word_embedding}.
On the other hand, visual attributes $\sps{A}$ are usually restricted to entities which we have seen before, and for which we have training images and learned models.
This allows us to propagate gradients to $\mathrm{v}(\sps{A})$ during training and optimize the attributes embeddings which yields improved performance (see model analysis in \secref{subsec:zeroshot}).

\paragraph{Limited training data}
Learning $\mathbf{R}$ directly from training data is not favorable since the number of class-attribute associations available for training are usually small.
For example, a typical data set consisting of 40 categories and 80 attributes yields around 1500 positive associations compared to tens or even hundreds of thousands of parameters in $\mathbf{R}$.
Hence, in order to avoid overfitting we build on the ideas of~\cite{Jenatton2012} and reduce the number of parameters that are required to be learned, by representing the relation operator $\mathbf{R}_j$ as a combination of $L$ latent factors:
\begin{equation}
\eqtopvspace
\mathbf{R}_j=\sum_{l=1}^L\alpha_l^j\mathbf{\Theta}_l, \quad \alpha^j\in \mathbb{R}^L \mbox{~~and~~} \mathbf{\Theta}_l \in \mathbb{R}^{d \times d},
\eqbottomvspace
\end{equation}
where $\alpha^j$ is a sparse vector used to weight the contributions of the rank one latent factors $\mathbf{\Theta}$.
Both $\alpha$ and $\mathbf{\Theta}$ are learned while minimizing \eqref{eq:min_log} and constraining \mbox{$\|\alpha^j\|_1 \leq \lambda$}.
The parameter $\lambda$ controls the sparsity of $\alpha$, and hence the extent to which latent factors are shared across relations.
Modeling $\mathbf{R}$ with latent factors has the benefit of allowing the learned relations to interact and exchange information through $\mathbf{\Theta}$ and hence improves the ability of the model to generalize.

\paragraph{Type of relations}
In order to train our model, we need to define the relations that link classes with the respective attributes.
Usually these relations are harvested through the process of collecting and annotating attributes (\eg what color is a bear? what shape is a bus?).
We refer to this type of relations as \emph{semantic relations}.
However, while some data sets do provide such relation annotations~\cite{Farhadi2009,Wah2011} others do not~\cite{Lampert2013}.
An alternative approach to manual annotation is to automatically discover relations by utilizing the word embedding space.
As described earlier in \secref{sec:app_word_embedding}, embeddings of semantically related entities tend to be close to each other (see \figref{fig:tsne_attr}).
Hence, one can simply group attributes into several relations by clustering their embeddings (\ie $N$ = number of clusters).
We refer to this type of relations as \emph{data driven relations}.


\subsection{Predicting binary associations}\label{sec:app_predicting_associations}
\makeatletter{}
Given an unseen class $z$, we predict its associations with the attribute set $\sps{A}$:
\begin{equation}
\eqtopvspace
r_j(z,a_m) = \left\{
  \begin{array}{l l}
    1 & \, \mathrm{if} \, p(r_j(z,a_m)) > t_+\\
    0 & \, \mathrm{if} \, p(r_j(z,a_m)) < t_-\\
    \varnothing & \, \mbox{otherwise}
  \end{array} \right. \quad \forall m,
\eqbottomvspace
\end{equation}
where thresholds $t_+$ and $t_-$ are learned to help select the most confident positive and negative associations while at the same time provide enough discriminative attributes to predict a novel class.
Assignment to $\varnothing$ discards the attribute for ZSL since we are not confident about the type (positive or negative) of the association.
We learn these thresholds using leave-K-class-out cross-validation so as to maximize zero-shot classification accuracy of the held out classes.

\paragraph{Zero-shot learning}
The score for unseen class $z$ on image $x$ is estimated based on the predicted attribute associations $(\sps{A}^z=\{a_m^z\} \subseteq \sps{A})$ using the Direct Attribute Prediction (DAP)~\cite{Lampert2009} method:
\begin{equation}
\eqtopvspace
s(z|x) = \prod_{a_m \in \sps{A}^z} p(a_m=a_m^z|x)/p(a_m),
\eqbottomvspace
\end{equation}
where $p(a_m|x)$ is the posterior probability of observing attribute $a_m$ in image $x$.
We assume identical class and attribute priors.

\section{Experiments}
\makeatletter{}

\label{sec:experiments}
In this section, we evaluate our model at: (1) predicting class-attribute associations and (2) unsupervised zero-shot classification.
Furthermore, we demonstrate the ability of our model to (3) transfer attributes across data sets without the cost of additional annotations.
Finally, (4) we show that the model is generic and can learn different types of relations and not only attribute-based ones.
In the following, we refer to our Class-Attribute Association Prediction model as \ours{}.

\paragraph{Data setup}
\label{subsec:datasetup}
We use two publicly available data sets.\\
(i) Animals with Attributes (AwA)~\cite{Lampert2009}: consists of 50 animal classes that are described with 85 attributes.
The classes are split into 40 seen and 10 unseen classes for ZSL.\\
(ii) aPascal/aYahoo (aPaY)~\cite{Farhadi2009}: contains 32 classes of artifacts, people and animals; and they are described with 64 attributes.
20 of these classes (aPascal) come from the Pascal challenge~\cite{pascal-voc-2008} and are used for training, while the rest 12 (aYahoo) are considered unseen and used for ZSL.

\subsection{Predicting class-attribute association}
\label{subsec:cls_attr_assoc}
\makeatletter{}
\begin{table}[!t]
\centering
\begin{tabular}{l c c}
\toprule
Model 						& AwA  			& aPaY 			\\
\midrule
Co-Occurrence~\cite{Mensink2014,Rohrbach2010} 				&				&				\\
\hspace{30pt} Bing			& 41.8 (57.4)	& 20.9 (69.4)	\\
\hspace{30pt} Yahoo-Img\footnotemark{}		& 50.9 (62.5)	& -	\\
\hspace{30pt} Flickr		& 48.7 (63.4)	& 28.1 (82.3)	\\
Word Embedding 				&				&				\\
\hspace{30pt} C $\rightarrow$ A (Top Q)			& 41.3 (53.7)	 & 34.2 (74.0)	\\
\hspace{30pt} C $\rightarrow$ A (Similarity)	& 41.3 (43.1)	 & 34.2 (77.5)	\\
\midrule
Ours &	&	\\
\hspace{30pt} \ours{}~(SR)			& 79.1 (78.2)	 & \textbf{76.1 (89.8)}	\\ \hspace{30pt} \ours{}~(DR)			& \textbf{79.7 (78.9)}	 & 75.7 (89.6)	\\ \bottomrule

\end{tabular}
\tablblvspace
\caption{Performance of class-attribute association predictions for unseen classes, presented in mAP (accuracy).}
\label{tab:pred_assoc}
\tabvspace
\end{table}
 
\makeatletter{}We consider two types of relations for training our \ours{} model:

\paragraph{Semantic relations (SR)}
For aPaY, we use the 3 predefined relations (\emph{has\_material}, \emph{has\_shape} and \emph{has\_part}). As for AwA, a cursory look at the set of attributes shows us that they can be easily grouped into 9 sets of relationships like \emph{has\_color}, \emph{lives\_in}, \emph{food\_type}, \etc.\footnotetext{\vspace{+0.02cm}We use Yahoo Image association scores provided by~\cite{Rohrbach2010} for AwA.}\footnote{\label{suppnote}More detailes can be found in the supplementary material.}

\paragraph{Data-driven relations (DR)}
For both data sets, we perform hierarchical agglomerative clustering on the word embeddings of the attributes and by analyzing the respective dendrogram the clustering is stopped at 10 groups of attributes.

\vspace{0.2cm}
We generate both positive and negative training triplets using the attribute annotations of the training set (\eg~$\mathrm{has\_part(horse, tail )=1}$, $\mathrm{lives\_in(dolphin, desert )=0}$).
We estimate the number of latent factors $L$ and $\lambda$ using 5-folds cross validation. We report the performance to predict all attribute associations for the unseen classes, hence we set $t_-=t_+=0.5$.
For words embedding, we train a skip-gram model on the Wikipedia corpus and obtain a $d=300$ dimensional representation.

We compare our method of predicting class-attribute associations via word vector representations and learned semantic relationships against the state-of-the-art (SOTA) Co-occurrence approach and two other baselines based on Word embedding space.

\paragraph{Co-occurrence}
As in the state-of-the-art methods~\cite{Mensink2014,Rohrbach2010}, we use the Microsoft Bing Search API~\cite{BingAPI}, the Flickr API~\cite{FlickrAPI} and Yahoo Image to obtain hit counts $H_{c_k}$ for classes (\eg~``chimpanzee''); $H_{a_m}$ for attributes (\eg~``stripes''), and $H_{c_k,a_m}$ jointly for class-attribute pairs (\eg~``chimpanzee stripes'').
We use the Dice score metric~\cite{Mensink2014} to obtain a hit-count based class-attribute association score:
\begin{equation}
\eqtopvspace
s^{H}_{c_k,a_m} = \frac{H_{c_k,a_m}}{H_{c_k} + H_{a_m}} \, ,
\eqbottomvspace
\end{equation}
where $s^{H}$ is the co-occurrence similarity matrix of classes and attributes.

\paragraph{Word embedding space}
These methods directly use the word vector representations (\secref{sec:app_word_embedding}) to predict class-attribute associations.
We present two approaches using the word embeddings:

\noindent (i) C $\rightarrow$ A (Top Q):
Consider the average number of attributes that are associated with every class in the training set to be Q.
For each unseen class, we consider a positive association with the Q nearest attributes (in terms of Euclidean distance) using the vector space embedding.

\noindent (ii) C $\rightarrow$ A (Similarity): Similar to the co-occurrence method, we construct a similarity matrix between class and attribute labels as:
\begin{equation}
\eqtopvspace
s^{W}_{c_k,a_m} = \exp(- \|\mathrm{v}(c_k) - \mathrm{v}(a_m)\|_2 ) \quad \forall c_k, a_m.
\eqbottomvspace
\end{equation}

For $s^{H}$ and $s^{W}$, binary associations are obtained by choosing the best threshold over the class-attribute similarity matrix which maximizes the ZSL performance.

\paragraph{Results}
\tblref{tab:pred_assoc} presents the mean average precision (mAP) and accuracy for predicting class-attribute associations. Note that among co-occurrence methods Flickr and Yahoo Image search perform better than Bing web search.
This can be related to the fact that the search results are grounded from visual information.
As demonstrated earlier, the word embedding space is not suitable to directly model the relations and it fails to reliably predict class-attribute associations (see Fig.~\ref{fig:tsne_cls-attr}).

\begin{figure}[!t]
\centering
\begin{subfigure}[b]{0.47\linewidth}
    \centering
    \includegraphics[height=\linewidth]{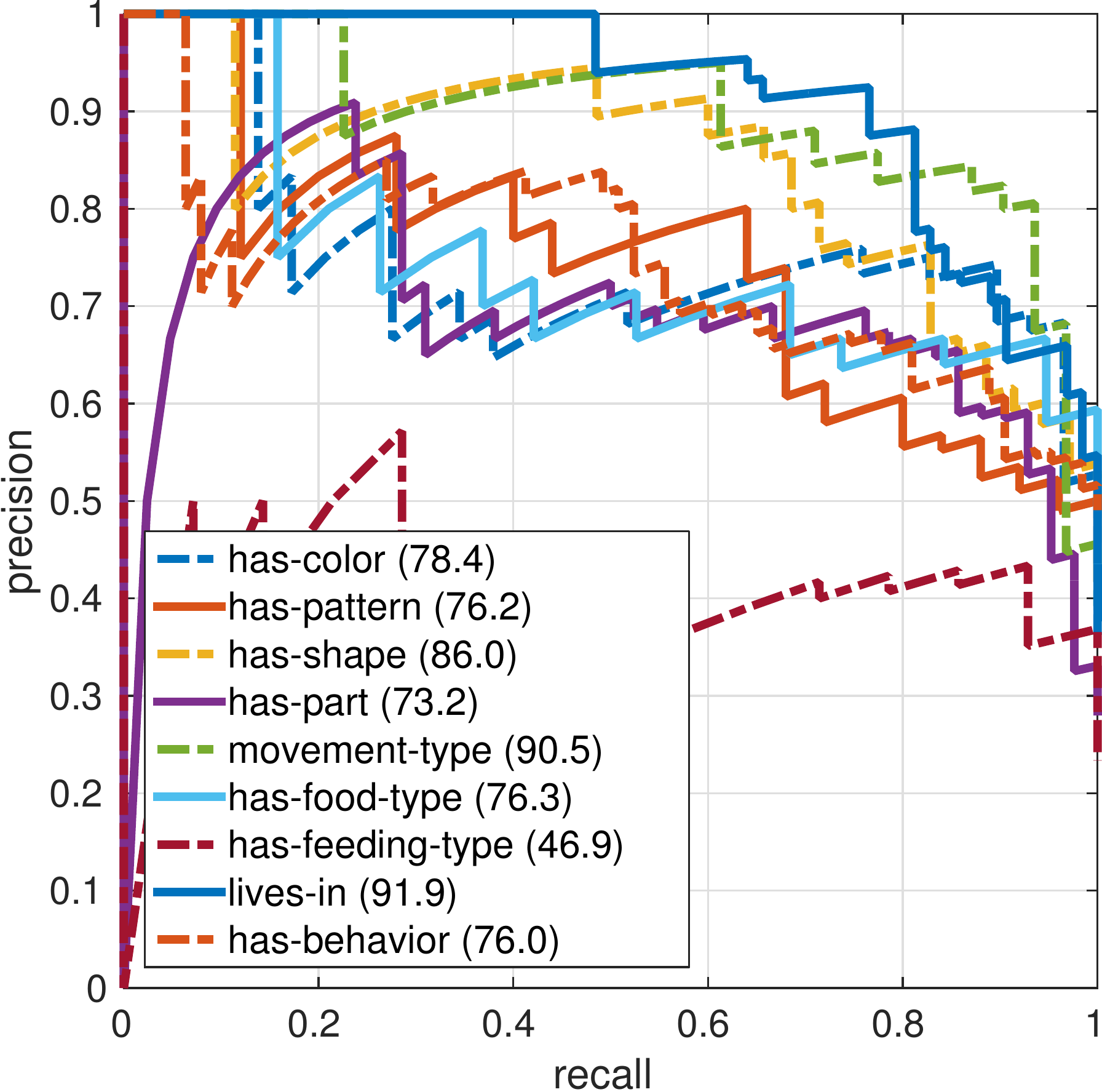}
    \caption{AwA}\label{fig:rel_pr_awa}
\end{subfigure}\quad
\begin{subfigure}[b]{0.47\linewidth}
    \centering
    \includegraphics[width=\linewidth]{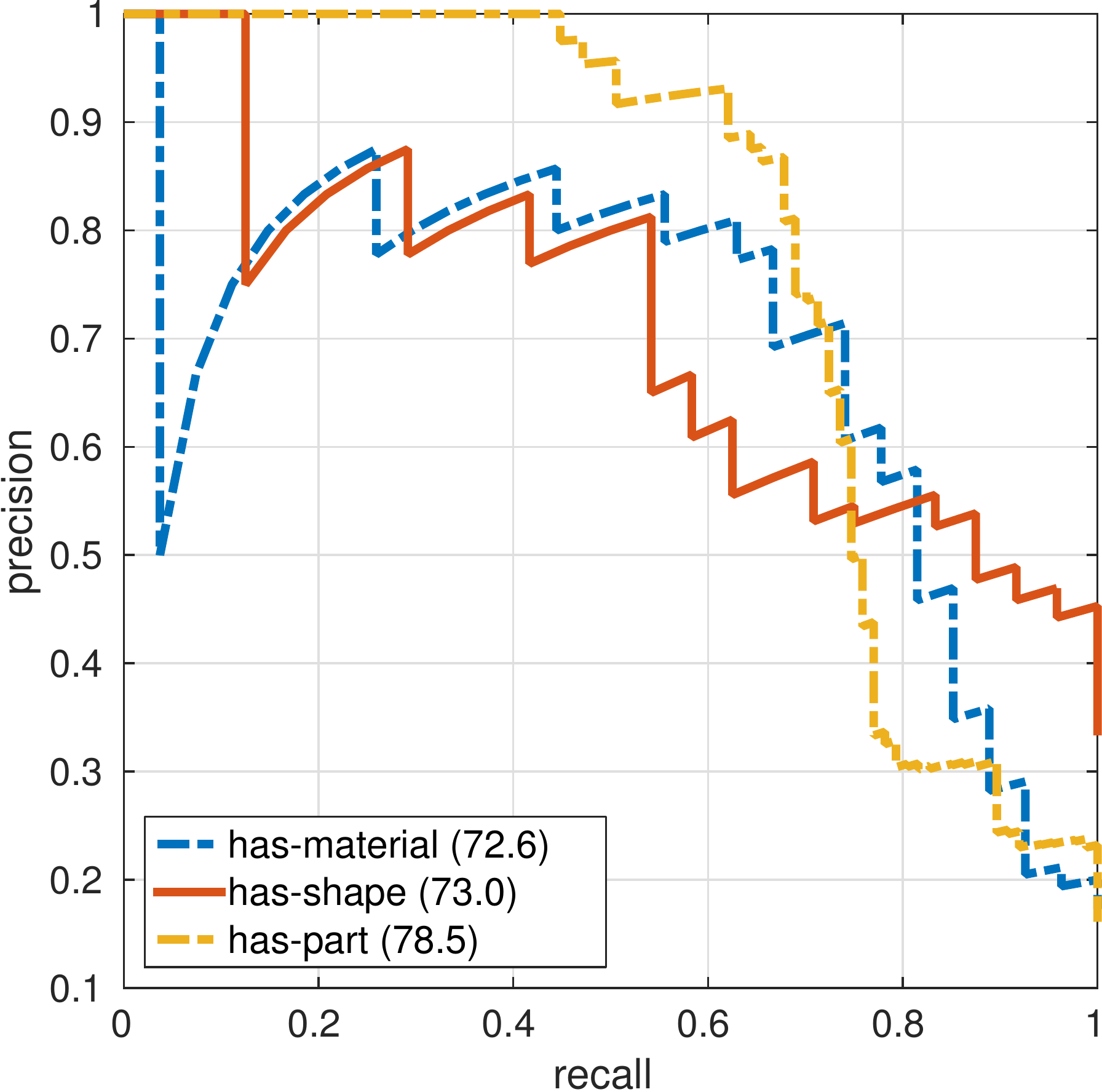}
    \caption{aPaY}\label{fig:rel_pr_apay}
\end{subfigure}
\figlblvspace
\caption{Prediciton performance of individual relations learned by \ours{} given by precision-recall curves along with mAP scores (see legend).}
\label{fig:assoc_pred_relations}
\figvspace
\end{figure}

Our method of modeling relations outperforms state-of-the-art by a significant amount (19\% on AwA, 42\% on aPaY).
\tblref{tab:pred_assoc_ex} presents examples of the top 5 confident positive and negative associations.
In general, we observe that our model ranks the most distinctive attributes of a category higher (\eg leopard$\leftrightarrow$fast, chimpanzee$\leftrightarrow$walk, hippopotamus$\leftrightarrow$strong).
\figref{fig:assoc_pred_relations} provides a deeper insight on the performance of each semantic relation presented by the precision-recall curve. 

Moreover, both SR and DR models perform at the same level with no substantial difference.
Hence, the data-driven approach is a very good alternative for the semantic relations thus even removing the need to provide extra relation annotations for \ours{}.
In the rest of the experiments we adopt the DR approach.
 
\makeatletter{}
 
\makeatletter{}
\newcommand{\fsz}{\fontsize{7.5pt}{7.0pt}\selectfont }
\newcommand{\mrk}[1]{{\color{gray}\textit{#1}}}
\begin{table}[!t]
\renewcommand{\arraystretch}{0.9}
\centering
\scalebox{1.0}{{\small
\begin{tabular}{l p{2.5cm} p{2.5cm}}
\toprule
				& \multicolumn{2}{c}{Top Associations}	\\
Unseen Category	& \multicolumn{1}{c}{Positive}	& \multicolumn{1}{c}{Negative}	\\
\midrule
hippopotamus 	& \fsz strong,  \mrk{group},  big,  walks,  ground  
				& \fsz claws,  flys,  red,  nocturnal,  weak  \\ 
leopard 		& \fsz fast,  lean,  oldworld,  active,  tail  
				& \fsz tusks,  water,  arctic,  plankton,  weak  \\ 
humpback\_whale & \fsz fast,  ocean,  water,  group,  fish  
				& \fsz red,  weak,  tunnels,  nocturnal,  plains  \\ 
seal 			& \fsz fast,  \mrk{meatteeth},  \mrk{bulbous},  big,  toughskin  
				& \fsz grazer,  tunnels,  longleg,  hooves,  longneck  \\ 
chimpanzee 		& \fsz walks,  group,  fast,  chewteeth,  active  
				& \fsz arctic,  flippers,  red,  plankton,  strainteeth  \\ 
								\bottomrule
\end{tabular}}}
\tablblvspace
\caption{Examples of predicted class-attribute associations for unseen classes. Wrong associations are highlighted in gray and italic. }
\label{tab:pred_assoc_ex}
\tabvspace
\end{table}

\subsection{Unsupervised zero-shot learning}
\label{subsec:zeroshot}
\makeatletter{}
We now present unsupervised zero-shot classification performance comparing against methods of the previous section which also use predicted class-attribute associations.
For all attribute-based approaches, we use the DAP model as described in \secref{sec:app_predicting_associations}.

\paragraph{Word embedding space}
In addition to the previous attribute-based baselines, we also examine two category-based options:

\noindent (i) C $\rightarrow$ C (Top 1):
As we see from Fig.~\ref{fig:tsne_cls}, similar classes do appear together in the word embedding space.
Hence, for each unseen class we use the category classifier of the training set class which appears closest to it in the vector space.

\noindent (ii) C $\rightarrow$ C (Weighted K):
This takes into consideration the similarity of the novel class for all known classes~\cite{Bart2005a}.
We build a classifier as a weighted linear combination of all training classes where the weights are based on distances between their vector representations:
\begin{equation}
\eqtopvspace
s_{wc}(z|x)=\sum_{k=1}^K w_k^zs(c_k|x),
\eqbottomvspace
\end{equation}
where $w_k^z=\exp(-\|\mathrm{v}(z) - \mathrm{v}(c_k)\|_2)$ and $s(c_k|x)$ is the score obtained by the classifier for category $c_k$ on image $x$.

Furthermore, most previous works assume that the attribute labels are provided by a human operator for the unseen class.
While in this paper we circumvent this additional overhead, we present the results of supervised DAP~\cite{Lampert2013} as a reference.

\paragraph{Image features and classifiers}
We use the output of the last hidden layer of the public BVLC implementation~\cite{Jia2014} of GoogLeNet~\cite{Szegedy2014} as our 1024 dimensional image features.
The deep representation is then used to train linear SVMs with regularized logistic regression~\cite{REF08a}  for the attribute and category classifiers.
The SVM parameter $C$ is estimated using 5-folds cross validation.
The same classifiers are used for the various baselines and our model. For our model we estimate the number of latent factors $L$ and $\lambda$ and additionally learn the thresholds ($t_-$, $t_+$) by 5-folds cross validation.

\paragraph{Results}
\tblref{tab:zero_shot} presents the mean per class accuracy for the test classes used in zero-shot classification.

We see again that image-based hit-count information obtained by Yahoo images or Flickr outperforms general web-based search (Bing).
However, they are all far from supervised ZSL performance with ground truth association.

The word embedding methods based on attributes (C $\rightarrow$ A) show poor performance.
In comparison, using the classifier of the nearest class (C $\rightarrow$ C (Top 1)) performs well for AwA (48\%) while poorly on aPaY (15\%).
An explanation for this is that unseen classes on AwA are visually close to the train set while aPaY has higher diversity in class types (animals and man-made objects).
Building a classifier by weighting all other classes (C $\rightarrow$ C (Weighted K)) shows moderate performance on both data sets. 

Our method outperforms all baselines with an accuracy of 67.5\% for AwA and 37.0\% for aPaY .
In fact, for aPaY \ours{} performs at the same level of supervised DAP while for AwA we see impressive performance surpassing the performance of supervised ZSL with ground truth attribute associations.
We show in \secref{subsec:share_attr} that our method allows to conveniently transfer attributes across data sets with no additional effort.
Using automatic transfer we can improve performance even more on both data sets.
\makeatletter{}
\begin{table}[!t]
\centering
\begin{tabular}{l c c}
\toprule
Model 						&	AwA 		&	aPaY	\\
\midrule
Supervised ZSL & & \\
DAP~\cite{Lampert2013}	&	59.5		&	37.1	\\
\midrule
Unsupervised ZSL & & \\
Co-Occurrence~\cite{Rohrbach2010,Mensink2014} &	& \\
\hspace{30pt} Bing			&	11.8		&	13.1	\\
\hspace{30pt} Yahoo-Img		&	39.8		&	-		\\
\hspace{30pt} Flickr		&	44.2		&	13.8	\\
Word Embedding 				&				&			\\
\hspace{30pt} C $\rightarrow$ A (Top Q)		&	10.2	&	14.3			\\
\hspace{30pt} C $\rightarrow$ A (Similarity)&	26.4	&	20.4			\\
\hspace{30pt} C $\rightarrow$ C (Top 1)		&	48.6	&	15.0			\\
\hspace{30pt} C $\rightarrow$ C (Weighted K)&	40.6	&	22.5			\\
\midrule
\ours{}~(ours)				& \textbf{67.5}& \textbf{37.0}	\\ \bottomrule

\end{tabular}
\tablblvspace
\caption{Zero-shot classification performance presented in mean per-class accuracy.}
\label{tab:zero_shot}
\tabvspace
\end{table}

\paragraph{Model analysis}
We study the effect of the different aspects of our model on the final unsupervised ZSL performance.

\noindent \textbf{(1)} Single relation: In the previous experiments, we used a small set of relations that group similar attributes together.  Here, we group all attributes in a single abstract relation ($N = 1$) called \emph{has\_attribute} and try to model the class-attribute associations accordingly.
We observe that in this setting, the absolute drop in accuracy is 5\% on AwA while on aPaY we see a reduction  by 12\%.

\noindent \textbf{(2)} Fixed attribute embedding: Similar to the category embeddings, we fix the representation of the attributes during learning.
Here the performance on both data sets drops by 2\% on AwA and 9\% on aPaY.

\noindent \textbf{(3)} Threshold@0.5: Rather than learning the thresholds $(t_-,t_+)$ we set them both to $t_-=t_+=0.5$.
In this case, the accuracy drops by 2\% on AwA while the performance on aPaY goes down by 12\%.

We conclude that improving the attribute representation during learning is beneficial.
We notice that attribute pairs like (\emph{big}, \emph{small}) and (\emph{weak}, \emph{strong}) which get initialized with similar embeddings are pushed apart by our model to facilitate the learning of the relations. 
It is also good to learn multiple relations that account for the discrepancies in the attributes rather than an abstract mapping that groups all of them together in one inhomogeneous group.
Our model learns proper confidence scores on the associations, and ranks most distinctive attributes higher leading to better ZSL performance when considering the most confident associations.
In general, aPaY is more sensitive to changes which can be related to the large variance in both classes and  attributes, since they describe not only animals (like in AwA) but also vehicles and other man made objects. 

\makeatletter{}
\begin{table}[!t]
\setlength{\tabcolsep}{4.0pt}
\centering
\scalebox{1.0}{\begin{tabular}{l c c c}
\toprule
					&	\multicolumn{3}{c}{\textbf{Target} (\emph{unseen})} \\
\textbf{Source} (\emph{seen}) 	&	AwA		&	aPaY		&	AwA+aPaY \\
\midrule
AwA 				&	{\color{gray}67.5}	&	39.5				&	37.1			\\
aPaY				&	10.4				&	{\color{gray}37.0}	&	~~6.2			\\
AwA+aPaY			&	\textbf{68.6}		&	\textbf{49.0}		&	\textbf{46.8}	\\
\bottomrule

\end{tabular} }
\tablblvspace
\caption{Zero-shot classification accuracy when attributes are transferred across data sets using \ours. A source AwA and Target aPaY means classifying the unseen classes of aPaY based on their predicted associations with the attributes of AwA. 
}
\label{tab:transfer_attr}
\tabvspace
\end{table}
 
\subsection{Attribute transfer across data sets}
\label{subsec:share_attr}
\makeatletter{}
A major advantage of our approach is the ability to automatically transfer the set of attributes from one data set to another at no additional annotation cost.
For example, we can use the 85 attributes of the AwA data set to describe categories from aPaY and vice-versa.
Most importantly, we do \emph{not} need to manually associate the classes of one data set with the attributes of the other.
These associations are automatically obtained through our \ours{} model.

In particular, we learn the relations of AwA and aPaY jointly without providing any additional associations.
Then for a novel class (from AwA or aPaY), we predict its associations to the attribute set $\sps{A}=\sps{A}^{AwA} \cup \sps{A}^{aPaY}$.
We see in \tblref{tab:transfer_attr} (3rd row), that \ours{} results in a significant improvement surpassing the performance of the manually defined associations on each data set.
Especially on aPaY, we see a dramatic improvement of 12\% in performance which can be attributed to the fact that roughly half the classes of aPaY test set are animals, which benefit strongly from the rich attributes transferred from AwA.
This demonstrates the effectiveness of \ours{} in integrating knowledge from different sources without the need for any additional effort.

In \tblref{tab:transfer_attr}, we provide additional evaluation of the attribute transfer by changing the source and target sets.
Comparing the two sources AwA and aPaY, it is clear that AwA encompasses a richer diversity as it results in good performance for both test sets, while transferring attributes from aPaY$\rightarrow$AwA results in performance on par with a random classifier.
Taking a closer look at the assigned attributes, we notice the following:

\noindent \textbf{(1)} AwA$\rightarrow$aPaY, not only the animal classes but even some man made classes get associated with reasonable attributes.
For example, the class ``jetski'' is positively associated with attributes ``water" and ``fast"; and class ``carriage" with ``grazer" and ``muscle".

\noindent \textbf{(2)} aPaY$\rightarrow$AwA, the attributes assigned to the animal classes are in general correct.
However, aPaY doesn't have enough animal-related attributes to distinguish the fine grained categories on AwA.
Most of the test classes in AwA are assigned to attributes like ``eye", ``head" and ``leg".

\noindent \textbf{(3)} AwA+aPaY$\rightarrow$AwA+aPaY, even on this harder setting where we test on 22 unseen classes (\ie~random performance drops to 4.5\% as compared to 10\% on AwA and 8.3\% on aPaY).
Our model generalizes gracefully with 46.8\% accuracy.

\subsection{\ours{} versus state-of-the-art}
In \tblref{tab:uzsl_sota}, the performance of our approach is compared against state-of-the-art in unsupervised ZSL.
Both \cite{Frome2013} and \cite{Norouzi2014} use the same word embedding as ours while \cite{Akata2015} uses GloVe \cite{Pennington2014} and Word2Vec.
Additionally, all methods in \tblref{tab:uzsl_sota} use image embedding from GoogLeNet.
\ours{} outperforms approaches based only on class names~\cite{Frome2013,Norouzi2014} with more than 20\% on both data sets.
Approaches like Text2Visual~\cite{Elhoseiny2013,Bo2010}, SJE~\cite{Akata2015} and HAT~\cite{Al-Halah2015} make use of additional source of information like Wikipedia articles or WordNet.
While theoretically this information can be obtained automatically, practically, a manual intervention is often necessary to resolve ambiguities between class names and article titles or to find the proper synset of a class in WordNet.
Nonetheless, \ours{} outperform state-of-the-art by 8.5\% and 18.8\% on AwA and aPaY respectively, while only needing the name of the unseen class.
 
\makeatletter{}
\begin{table}[!t]
\centering
\begin{tabular}{l l c c}

\toprule
Model 						& ZSL Information 	&	AwA 		&	aPaY	\\
\midrule
DeViSE~\cite{Frome2013} 	& C					&	44.5		&	25.5	\\
Text2Visual~\cite{Elhoseiny2013,Bo2010}	& C + $\mathrm{Text}^\mathrm{Wiki}$	
												& 	55.3		&	30.2	\\
ConSE~\cite{Norouzi2014}	& C					&	46.1		&	22.0	\\			
SJE~\cite{Akata2015}		& C + $\mathrm{H}^\mathrm{WordNet}$	
												&	60.1		&	-		\\
HAT~\cite{Al-Halah2015}\footnotemark{}		& C + $\mathrm{H}^\mathrm{WordNet}$
												&	59.7		&	31.1	\\
\midrule
\ours{}~(ours)				& C 				& \textbf{68.6}	& \textbf{49.0}	\\
\bottomrule

\end{tabular}
\tablblvspace
\caption{Unsupervised zero-shot learning accuracy of state-of-the-art versus \ours{}. The second column shows the type of information leveraged by each model for the unseen classes.}
\label{tab:uzsl_sota}
\tabvspace
\end{table}

\subsection{Beyond attributes}
\label{subsec:beyond_attr}
\makeatletter{}Various approaches in the literature have reported the advantage of incorporating hierarchical information for ZSL (\eg~\cite{Al-Halah2015,Akata2015,Rohrbach2011}).
Our model can also learn hierarchical relations, for example to predict the ancestors of a category.
To test this, we query WordNet~\cite{Miller1995} with the AwA categories and extract the respective graph relevant to the hypernym links.
We then learn the \emph{has\_ancestor} relation by generating triplets of the form $\mathrm{has\_ancestor(horse,equine)}=1$ using the information from the extracted graph.
The evaluation on AwA test set reveals that we can predict the ancestor relation of an unseen class with a mAP of 89.8\%.
Interestingly, learning such a hierarchy-based relation can aid the learning of some attribute-based relations.
The model allows the various relations to interact and exchange information at the level of the shared latent factors.
Among the improved attribute-based relations, is has\_pattern (+2.5\%), and feeding\_type (+2.1\%).
These relations correlate well with the hierarchical information of the classes (\eg~carnivores tend to have similar pattern and feeding type).
Predicting such a hierarchical relation alleviates the need of a complete hierarchy or manual synonym matching since this is automatically handled by the word embedding and \ours{} model.
This keeps user intervention to the minimal requirement of providing class names.
We expect that modeling more relations among the classes jointly with class-attribute relations can result in better performance.

\footnotetext{Results are from the updated arXiv version of \cite{Al-Halah2015}: \href{http://arxiv.org/abs/1604.00326}{1604.00326v1}}
\section{Conclusion}
\makeatletter{}
Attribute-based ZSL suffers from a major drawback of needing class-attribute associations to be defined manually.
To counter this, we present an automatic approach to predict the associations between attributes and unseen classes.
We model the associations using a set of relationships linking categories and their respective attributes in an embedding space.
Our approach effectively predicts the associations of novel categories and outperforms state-of-the-art in two tasks; namely association prediction and unsupervised ZSL.
Moreover, we demonstrate the ability of our model to transfer attributes between data sets at no cost.
The transferred attributes enlarge the size of the description vocabulary, which results in more discriminative classifiers for ZSL yielding an additional boost in performance.

\balance
{\reffontsize
\bibliographystyle{ieee}
\bibliography{cvpr16}
}

\end{document}